# ActKnow: Active External Knowledge Infusion Learning for Question Answering in Low Data Regime

**Annervaz K M**
Computer Science and Automation
Indian Institute of Science
annervaz@iisc.ac.in

**Pritam Kumar Nath**
Computer Science and Automation
Indian Institute of Science
pritamnath@iisc.ac.in

**Ambedkar Dukkipati**
Computer Science and Automation
Indian Institute of Science
ambedkar@iisc.ac.in

## Abstract

Deep learning models have set benchmark results in various Natural Language Processing tasks. However, these models require an enormous amount of training data, which is infeasible in many practical problems. While various techniques like domain adaptation, fewshot learning techniques address this problem, we introduce a new technique of actively infusing external knowledge into learning to solve low data regime problems. While it has been shown that infusion of external knowledge can improve the performance of models, till now, there are no studies on its usefulness in solving low-data regime problems. We propose a technique called ActKnow that actively infuses knowledge from Knowledge Graphs (KG) based "on-demand" into learning for Question Answering (QA). By infusing world knowledge from Concept-Net, we show significant improvements on the ARC Challenge-set benchmark over purely text-based transformer models like RoBERTa in the low data regime. For example, by using only 20% training examples, we demonstrate a 4 % improvement in the accuracy for both ARC-challenge and OpenBookQA, respectively.

## 1 Introduction

Machine learning problems in low data regimes arise naturally in many applications, as abundant labeled data may not be feasible. For example, collecting tens of thousands of question-answer pairs to build a Question Answering (QA) (Rajpurkar et al., 2016) system catering to a specific domain requires an enormous amount of resources and manpower. Most state-of-the-art QA systems do not do well in low data regime settings. In this setting, they lack the ability to learn commonsense dependencies hence restricted to dependencies learnable from training data. In this paper, we address the following. *Can low data regime problems be solved by infusing external knowledge into learning algorithms? Can this be done 'actively' or 'on-demand'?*

The models for open-domain question answering (refer to Table 1) ranges over large pretrained models like BERT (Devlin et al., 2018), RoBERTa (Liu et al., 2019), T5 (Colin Raffel, 23 Oct 2019) to models that use external data in various forms to achieve better performance. Models like UnifiedQA (Daniel Khashabi, 2 May 2020) uses 17 QA datasets spanning 4 diverse formats to build a unified question answering model, while others like AristoBERTv7 (Peter Clark, 2 Feb 2021) use a pretrained BERT (Devlin et al., 2018) model and finetune it on an external dataset like RACE (Lai et al., 2017a). MultitaskBERT (Xiaodong Liu, 31 Jan 2019) leverages large amounts of cross-task natural language understanding data and benefits from a regularization effect that leads to more general representations in order to adapt to new tasks and domains. A careful selection of knowledge ,Pratyay Banerjee (24 Jul 2019) is based on abductive information retrieval (IR) and information gain-based re-ranking for the OpenBookQA dataset (Mihaylov et al., 2018). Several of these models use external knowledge in the form of Knowledge Graphs like Improving QA with External Knowledge (Pan et al., 2019a) which uses both open-domain Knowledge Graphs and in-domain text enrichment. This paper proposes an active knowledge infusion framework for QA that uses GCN embedding, entity-relationship attention, and pretrained language models.

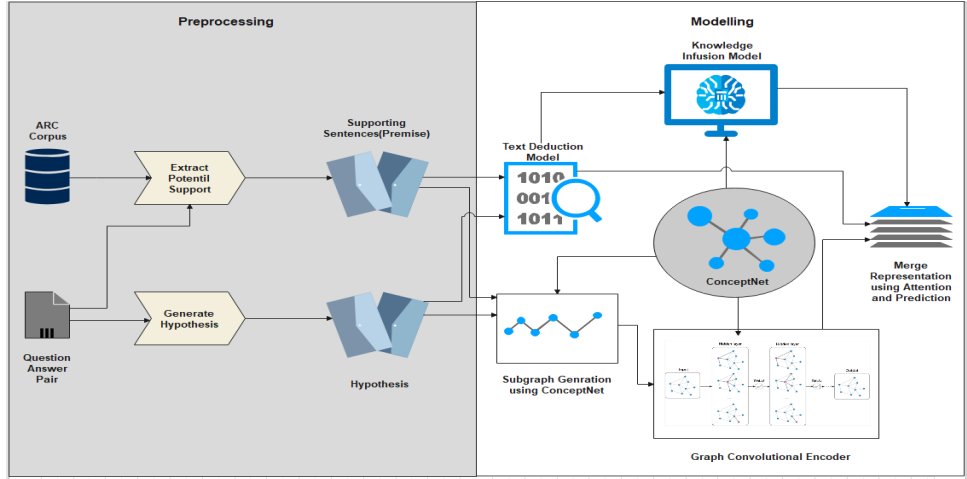

Figure 1: Schematic depiction of our approach. Text Encoding, Graph Encoding, and External Knowledge are incorporated in the model

Most QA models make a semantic embedding of a question and answer pair, assuming a flat structure to the text, neglecting complex dependencies between the tokens. If a KG from the domain is available, the tokens can be linked to the KG nodes. The sub-graph generated so from the KG acts as a source of dependencies between the tokens. However, the model may need external information about nodes not linked directly for deducing the answer. To incorporate these requirements, we propose a model with three main components, one that processes the text, the second one that creates a sub-graph from the KG about tokens present in the text, and the third one that fetches required external information from the KG. The active infusion of external information gained from KG becomes more important in a low data regime. The system has only a handful of training examples to learn from and has low confidence about the choice it makes. To address this, we propose an end-to-end trainable strategy that boosts the model performance in the low data regime.

**Contributions:**
**(1)** End-to-end trainable model for QA, incorporating the textual information, complex dependencies of the tokens, and relevant external knowledge. Using Graph Convolutional Network (GCN) (Kipf and Welling, 2016) based encoder with an attention-based method to integrate all three relevant information. We refer to this model as BASEKNOW. **(2)** Based on the above model, we propose ACTKNOW, which *actively* infuses external knowledge from KGs on demand. We develop a mechanism based on the entropy of softmax output that provides uncertainty of model for answering each sample question. **(3)** We use various training settings of ARC (Clark et al., 2018) and OpenBookQA (Mihaylov et al., 2018) dataset to show the significant improvement in accuracy, *especially in low data regime*s. For example, when using only 20% of OpenBookQA training examples, we show an improvement of 4% accuracy.

## 2 QA and Natural Language Inference

We follow the approach of converting Question Answering to that of Natural Language Inference (NLI) problem (Van Benthem et al, 2008; MacCartney and Manning, 2009). We briefly describe it here. Many QA systems share a number of characteristics with the NLI systems, and both problems require a good language understanding model. A multiple Choice Question Answering (MCQA) problem can be converted to Natural Language Inference (NLI) problem based on the natural language inference model predictions, the answer to the original question can be arrived at. Consider a question $Q$ and for potential answers given $A, B, C, D$, based on a text corpus. For a potential answer, say $A$, collect a certain set of sentences from the text corpus where $A$ appears using information retrieval techniques like elastic search (Gormley and Tong, 2015). Let $\mathcal{P}_A$ denote this set of sentences. In the question, replace the 'WH' word with the potential answer. Let us call this statement $\mathcal{H}_A$. Now treat $\mathcal{P}_A$ as the premise and $\mathcal{H}_A$ as the hypothesis, using the inference model check for entailment. The entailment implies that $A$ is the answer to the original question $Q$, and contradiction implies it is not the answer. The same holds for the other answer options. Many QA benchmarks such as SciTail (Khot et al., 2018; Zhang et al., 2018) are reoriented to natural language inference datasets. Here the hypothesis is created using the stemmed version of the question and answer choice, whereas the premise is supporting fact extracted from the supplementary information. The answer that generates a higher entailment score is the most feasible accurate answer. We employ this textual transformation to utilize the entailment prediction models in our approach for question answering. Table 2 (see section 5) depicts this transformation with the help of two examples.

| Model | ARC-Easy set Accuracy | ARC-Challenge set Accuracy | OpenBookQA Accuracy |
|---|---|---|---|
| KG$^2$ (Zhang et al., 2018) | — | 31.70 | — |
| TriAN + BERT+ $f_{cs}^{air}$ + $f_{cs}^{ind}$ (Zhong et al., 2018) | — | 36.55 | 72.8 |
| Reading strategies (Sun et al., 2019) | 68.9 | 42.3 | 55.8 |
| Improving QA with External Knowledge (Pan et al., 2019b) | 73.2 | 44.8 | 65.0 |
| QA Transfer | 76.47 | 53.84 | 69.6 |
| AristroBERTv7 (Peter Clark, 2 Feb 2021) | 79.29 | 57.76 | 72 |
| BERT MultiTask (Xiaodong Liu, 31 Jan 2019) | 72.43 | 48.29 | 63.8 |
| Careful Selection of Knowledge (Pratyay Banerjee, 24 Jul 2019) | - | - | 72 |
| UnifiedQA(BART-uncased-large) (Daniel Khashabi, 2 May 2020) | **82.6** | 54.95 | 67.8 |
| RoBERTa $^a$ + GCN + ER attention (ACTKNOW) (ours) | 81 | **63** | **78** |

$^a$RoBERTa-large was finetuned on the RACE (Lai et al., 2017b) dataset

Table 1: Our model shows 3.83 and 5.4 % improvements over the benchmark model in the ARC-challenge set and OpenBookQA respectively.

## 3 Our Model

We follow (Van Benthem et al., 2008; MacCartney and Manning, 2009) for converting QA to that of Natural Language Inference (NLI) problem. Consider a question $Q$, and four potential answers are given $A, B, C, D$, based on a text corpus. For a potential answer, say $A$, collect a certain set of sentences from the text corpus where $A$ appears using information retrieval techniques like elastic search (Gormley and Tong, 2015). Let $\mathcal{P}_A$ denote these set of sentences. In the question, replace the 'WH' word with the potential answer. Let us call this statement $\mathcal{H}_A$. Now treat $\mathcal{P}_A$ as the premise and $\mathcal{H}_A$ as the hypothesis, using the NLI model check for entailment. The entailment implies that $A$ is the answer to the original question $Q$, and contradiction implies otherwise. A schematic diagram of our approach is depicted in Figure 1. After converting the QA to the NLI realm, the hypothesis and premise are processed separately. First, the hypothesis and premise are encoded using a plain text encoder. Then text hypothesis and text premise are converted to a graph form encoded using GCN to generate graph encodings. The text encoding generates attention over KG facts and fetches attention to weighted facts from the KG. Consider $t_i^j, g_i^j, \mathcal{G}_i^j$ as the representation generated for $i^{th}$ answer choice of $j^{th}$ question by the text encoding, graph encoding and external knowledge infusion respectively. The aggregated vector $\mathcal{A}_i^j = t_i^j : g_i^j : \mathcal{G}_i^j$ is then used to do classification. loss = $L_{CE}(\text{groundtruth}, \text{softmax}(W^T \mathcal{A}_i^j))$ where $W$ are the model parameters and $L_{CE}(,)$ denotes the cross-entropy loss. We minimize this loss averaged across the training samples to learn the various model parameters using stochastic gradient descent (Bottou, 2012).

### 3.1 Encoding Text (Generating $t_i^j$):

We experimented with various transformer-based models, RoBERTa (Liu et al., 2019) gave the best results in combination. Consider $t_i^j$ as the representation generated for $i^{th}$ answer choice of $j^{th}$ question by any of these text encoders.

### 3.2 Encoding Graph (Generating $g_i^j$)

Encoding a graph involves two steps, sub-graph generation, and subsequent sub-graph encoding. We use Concept-Net (Liu and Singh, 2004) as our structural knowledge base to create sub-graphs for each premise-hypothesis pair $(P, H)$. First, we identify concepts in each premise-hypothesis pair and then connect recognized concepts using a path in the KG identified by the depth-first search. Extracting contextually related knowledge from a noisy resource is still an open research problem. We restricted path length because we observed that increasing path length includes noisy concepts. A study on this is given in Figure 4a. We use GCN (Kipf and Welling, 2016) as the main component of the encoder. Each node in the sub-graph $C_g$ is initialized by a 300-dimensional Concept-Net Number batch word embedding. The nodes in subsequent hidden layers are updated as

$$f(H^\ell, C_g) = \sigma \left( \hat{D}^{-\frac{1}{2}} \hat{C}_g \hat{D}^{-\frac{1}{2}} H^\ell W^\ell \right),$$

where $\hat{C}_g = C_g + I$, $\hat{D}$ is degree matrix of $\hat{C}_g$ and $I$ represents the identity matrix. We get $G \in \mathbb{R}^{N \times D}$ as the output of GCN Encoder for a hypothesis-premise pair, here $N$ is the number of nodes in the graph and $D$ is the hidden layer's output size. Consider $G_i^j$ as the representation generated for $i^{th}$ answer choice of $j^{th}$ question by this graph-based encoding. Let $G_i^j = [G_{i1}^j, G_{i2}^j, \ldots, G_{in}^j]$ denotes GCN output and $t \in \mathbb{R}^d$ as text encoding model output. We apply attention based system between output of GCN encoder and text encoding model to generate a feature representation for graph based encoder as follows:

$$\alpha_{G_{ik}^j} = \frac{\exp(tG_{ik}^j)}{\sum_{K=1}^{|G_i^j|} \exp(tG_{iK}^j)}, \text{ and } g_i^j = \sum_{k=1}^{|G_i^j|} \alpha_{G_{ik}^j} G_{ik}^j$$

**Algorithm 1:** BaseKnow

**Input** : $\{\{\mathcal{H}_i^j, \mathcal{P}_i^j\}_{i=1}^4\}_{j=1}^n, C = Concept - Net$
**Output:** Model Parameters $(W)$

1 Loss = 0, W = Random()
   for $j = 1$ to $n$ do
2     for $i = 1$ to $4$ do
3         $t_i^j$ =RoBERTa($\mathcal{H}_i^j, \mathcal{P}_i^j$)
4                                                                                                                 // text encoding
5         $A, E$ =Subgraph-Gen($C, H_i^j, \mathcal{P}_i^j$)
6         /* A is the adjacency matrix and E is the number batch embedding                          */
7         $g_i^j$ =GCN-Encoder($A, E, t_i^j$)
8         $\mathcal{G}_i^j$ =KG-Infusion($C, t_i^j$)
9         $\mathcal{A}_i^j = t_i^j : g_i^j : \mathcal{G}_i^j$
10                                                                                                          // concat the embeddings
11        Loss+ = $L_{CE}$(groundtruth, softmax($W^T \mathcal{A}_i^j$))
12        /* Updating the functional form of loss                                                       */
13 for $epoch = 1$ to $NumOfEpochs$ do
14     Update-Model-Parameters($W$, Loss)
15     return $W$

### 3.3 External Knowledge Infusion: Generating $\mathcal{G}_i^j$ (BaseKnow)

In graph encoding model explained previously, only the concepts present in the hypothesis and premise are considered. However for inference a lot more concepts, not directly mentioned in the text, from the full KG might be required. This part of the model takes this into account. We try to identify important concepts(also referred as entities) and relations in our context which may assist in answering the question with more precision. We apply a attention based model over the whole entity and relation set. Let $\{e^{(i)}\}_{i=1}^{|E|}, e_i \in \mathbb{R}^d$ and $\{r^{(i)}\}_{i=1}^{|R|}, r_i \in \mathbb{R}^d$ be the list of Concept-Net entity and relation represented as their $d$ dimensional RESCAL (Quan Wang and Guo, 2017) embedding , $|E|$ and $|R|$ are number of entities and relations in Concept-Net. We attend over both the lists separately using the context vector $t \in \mathbb{R}^d$ obtained from the text encoding model as follows:

$$\alpha_{e_i} = \frac{\exp(te_i)}{\sum_{j=1}^{|E|} \exp(te_j)}, \quad \alpha_{r_i} = \frac{\exp(tr_i)}{\sum_{j=1}^{|R|} \exp(tr_j)}$$

The value of $|E|$ ranges in millions that might give very small attention weights for all the entities. To mitigate this, we used Gumbel Softmax (Jang et al., 2016) to generate $\alpha_{e_i}$, which discretizes the distribution over $\alpha's$. This will help in assigning higher scores only to more relevant entities while other entities get suppressed. We generate a feature representation for entity and relation as follows:

$$e = \sum_{i=1}^{|E|} \alpha_{e_i} e_i \text{ and } r = \sum_{i=1}^{|R|} \alpha_{r_i} r_i$$

---

**Algorithm 2:** ActKnow

**Input** : $\{\{\mathcal{H}_i^j, \mathcal{P}_i^j\}_{i=1}^4\}_{j=1}^n, C = Concept-Net$
**Output:** Model Parameters ($W$)

1  $W = Random()$
   **for** $epoch = 1$ to $NumOfEpochs$ **do**
2      Loss = 0
3      /* New loss function formulated for every parent training epoch                                */
4      **for** $j = 1$ to $n$ **do**
5          **for** $i = 1$ to $4$ **do**
6              $t_i^j =$ RoBERTa($\mathcal{H}_i^j, \mathcal{P}_i^j$)
7              $A, E =$ Subgraph-Gen($C, \mathrm{H}_i^j, \mathcal{P}_i^j$)
8              $g_i^j =$ GCN-Encoder($A, E, t_i^j$)
9              $\mathcal{G}_i^j =$ KG-Infusion($C, t_i^j$)
10             $\mathcal{E}_j = Entropy\ Of\ Question\ j$
11             /* if p is the confidence score of one answer, -1 * sum of p * log p across all answers   */
12             $\alpha_i^j = \mathcal{E}_j . g_i^j$ , $\beta_i^j = \mathcal{E}_j . \mathcal{G}_i^j$
13             $\mathcal{A}_i^j = t_i^j : \alpha_i^j : \beta_i^j$
14                                                                  // weighted concatenation of embeddings
15             Loss$+ = L_{CE}($groundtruth$,$softmax$(W^T \mathcal{A}_i^j))$ /* Updating the new functional form of loss for this parent epoch   */
16     **for** $subEpoch = 1$ to $NumOfSubEpochs$ **do**
17         /* For new loss function formulated in every parent epoch, training is done for a number of sub epochs   */
18         Update-Model-Parameters($W$, Loss)
19  **return** $W$

---

### 3.4 Active External Knowledge Infusion from KG (ActKnow)

Here we introduce an active mechanism that actively infuses knowledge from KG on demand. The hypothesis is that samples for which model cannot predict answers with certainty (or without confusion) require more external knowledge. In the active learning setup, the samples for which the models are more confused are labeled in priority by humans and introduced back into the training (Aggarwal et al., 2014). Hence one of our main idea in this paper translates to *replacing humans in the loop with KGs*.

We propose to measure the above-mentioned uncertainty of the model on a particular sample by Shannon's entropy. Suppose for $i^{th}$ choice of question $j$, the softmax output is $p_i(j) =$ softmax$(W^T \mathcal{A}_i^j)$. Now the confidence score of the model for $j^{th}$ question is measured using entropy functional as $S(j) = -\sum_i p_i(j) \log p_i(j)$. Higher values of $S(j)$ indicate that the model is confused in answering question $j$, and lower values indicate otherwise. With this, the weight-age of the external knowledge infusion in the subsequent training iteration for question $j$ is made proportional to $S(j)$. If the confidence score of predicting a single answer is close to one or zero, the entropy value will be close to zero, and thus the contribution from an external source will be neglected; otherwise, if they are close to equal distribution, the entropy value will be high; thus the contribution from external knowledge source will be high. We show that this training scheme, which actively infuses external knowledge in the learning process, gives a significant performance boost, especially in the low data regime. An algorithmic description of this is given in Algorithm 2.

## 4 Implementation & Training details

We used PyTorch (Paszke et al., 2017) implementation of (Radford et al., 2018)[1] as our 12-layered transformer-based text model. We built our graph-based models on top of this implementation. We pre-trained the graph deduction and external data infusion model for few epochs before training them combined with transformer models to ensure proper flow of gradients in these components. The models were trained using Adam's optimizer (Kingma and Ba, 2015) in a stochastic gradient descent (Bottou, 2012) fashion. We used batch normalization (Ioffe and Szegedy, 2015) while training. The final hyper-parameters used are given in table 6. The experiments were carried out on Tesla V100 GPU with 32GB VRAM, and preprocessing needs 128GB RAM. For one master epoch, the time taken is around 10 hours for full training data.

| Set | Question-Answer | Snippet of support sentence corresponding to each choice (Premise for each choice) | Hypothesis |
|---|---|---|---|
| Easy | Q. The movement of soil by wind or water is called ? A1. Condensation A2. Evaporation **A3. Erosion** A4. Friction | Q-A1. The condensed water is called dew. The water that is removed is called condensate.... Q-A2. The water used by a crop and evaporated from the soil is called evaporation, or ET.... Q-A3. When soil is washed away by water and wind, it's called soil erosion. Erosion of soil is caused by water, wind, ice, and movement in response to gravity.... Q-A4. Waves are caused by the frictional drag of the wind over the surface of the water. Wind-driven waves,.... | Q-A1. The movement of soil by wind or water is called Condensation Q-A2. The movement of soil by wind or water is called Evaporation Q-A3. The movement of soil by wind or water is called Erosion Q-A4. The movement of soil by wind or water is called Friction |
| Challenge | Q. A goat gets energy from the grass it eats. Where does the grass get its energy? A1. soil **A2. sunlight** A3. water A4. air | Q-A1. This mushroom gets its energy from decaying organisms in the soil. Plants get their energy from the soil and from the Sun.... Q-A2. Animals eat plants to get their energy, and the plants get their energy by absorbing sunlight.... Q-A3. The nestling gets water and energy from fat. Most water gets energy.... Q-A4. Producers, such as plants and algae, get their energy directly from sunlight, and get the materials they need from air and water.... | Q-A1. A goat gets energy from the grass it eats. soil does the grass get its energy Q-A2. A goat gets energy from the grass it eats. sunlight does the grass get its energy Q-A3. A goat gets energy from the grass it eats. water does the grass get its energy Q-A4. A goat gets energy from the grass it eats. air does the grass get its energy |

Table 2: Examples from ARC Easy and Challenge set. Second column has question along with answer choices. Correct answer is one in bold. Third column are few supporting sentences for each answer choice(also referred as premise). Fourth column is hypothesis formed using the stemmed version of question-answer choices

## 5 Datasets & Preprocessing

ARC Dataset and Concept-Net: The AI2 Reasoning Challenge (ARC) has been introduced in (Clark et al., 2018) for the QA task. Many neural state-of-the-art models for well-known SQuAD (Rajpurkar et al., 2016) and SNLI tasks performed only somewhat better than the random baseline. It contains questions created for human exams from natural science-related questions. The ARC corpus is also provided with the dataset that is partitioned into two sets: (i) ARC-Easy set consisting of questions whose direct evidences can be found in the ARC corpus, and (ii) ARC-Challenge set, on the other hand, comprises difficult questions that do not have such direct evidence and are wrongly answered by both word co-occurrence based and retrieval based algorithms. To identify the difference between the two sets, consider two examples shown in 2. For the first example in the table, it is easily noticeable that the premise for correct choice `soil`, contains a direct reference to the answer. On the other hand, for second example, model should infer that `goat` is an `animal` and `grass` is related to `plants` in some sense. This is where our Graph encoding model can learn to align and improve predictions. The separation of the Challenge set is done with the motive of encouraging work with difficult questions. Our work is focused primarily on the ARC-Challenge set, but our proposed model shows improvement on both sets.

The ARC question set $Q = \{q_i, a_i^{(1)}, ..., a_i^{(m)}\}_{i=1}^N$ contains $N$ science-related questions $q_i$ targeting students of age 8 through 13. Each question $q_i$ has $m$ answer choices, one of which is correct. This is also equipped with the ARC corpus consisting of 14M science-related sentences. These questions should be relatively easier to solve by any human possessing commonsense knowledge about basic concepts involved in the question and answer choices. This process inspires our model. We propose a novel general framework using both pieces of knowledge learned from textual sources and external graph-based sources to replicate human reasoning.

---
[1] `github.com/huggingface/pytorch-openai-transformer-lm`

| Set | Train | Dev | Test |
|---|---|---|---|
| **ARC-Easy** | 2251 | 570 | 2376 |
| **ARC-Challenge** | 1119 | 299 | 1172 |
| **OpenBookQA** | 4957 | 500 | 500 |

Table 3: Number of examples in train, dev and test set of ARC-easy, ARC-challenge and OpenBookQA dataset.

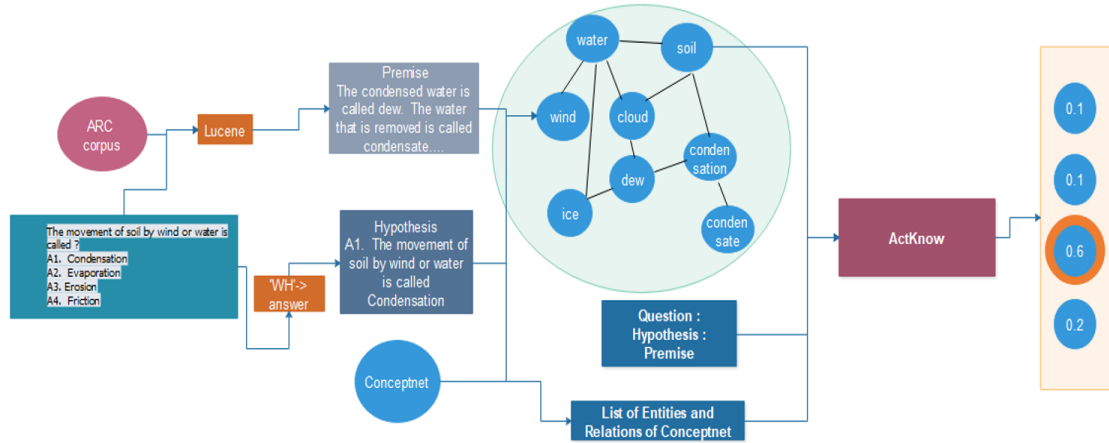

Figure 2: Data Preprocessing Pipeline

OpenBookQA: (Mihaylov et al., 2018) dataset consists of 5,957 multiple-choice elementary-level science questions (4,957 train, 500 dev, 500 test), which probe the understanding of a small "book" of 1,326 core science facts and the application of these facts to novel situations. For training, the dataset includes a mapping from each question to the core science fact it was designed to probe. Answering OpenBookQA questions requires additional broad common knowledge not contained in the book.

Generating Supporting facts, Hypothesis and Premise: Our approach depends upon the extraction of supporting facts as a premise for each answer option from the ARC corpus (Figure 2). Supporting facts are sentences present in the corpus that are likely to contain relevant information for a candidate's answer to a question being correct. For this purpose, we used a local search engine built on top of elastic search (Gormley and Tong, 2015) [2]. A hypothesis is created by combining question stem and answer choice using simple heuristics, such as replacing WH words in question stem with an answer option. Same transformations are also used by $KG^2$ (Zhang et al., 2018) for their inference models. For question $q$ with $m$ number of answer candidates, we have $m$ pairs of premise-hypothesis pairs.

## 6 Experiments and Result Discussion

We demonstrate the effectiveness of the proposed methods on the datasets ARC-Easy, ARC- Challenge (Clark et al., 2018) and OpenBookQA (Mihaylov et al., 2018). To demonstrate the effectiveness of the external knowledge fusion, we do not recourse to domain-dependent techniques like data augmentation, pre-training transformer model on RACE (Lai et al., 2017b) and SWAG ,(Zellers et al., 2018) etc. Our aim is to show that the text-based encoding model can be improved by introducing graph-based encoding and external knowledge infusion. Hence the design of our experiments is motivated by this. We have shown the comparison with the benchmark models in table 1. The best performing models (on the leaderboard[3],[4]) of the ARC and OpenBookQA dataset use very specific training strategy such as increasing dataset by combining different multiple-choice science QA datasets, pre-training transformer model on RACE (Lai et al., 2017b) and SWAG (Zellers et al., 2018) datasets or ensemble of Transformer model, etc. We argue that these training strategies are very domain-specific and might not generalize well to various NLP tasks across all domains.

---
[2] github.com/allenai/ARC-Solvers
[3] leaderboard.allenai.org/arc/submissions/public
[4] leaderboard.allenai.org/open_book_qa/submissions/public

| Text Encoding model | External Knowledge Infusion Model | Graph Encoding model | ARC-Easy set Accuracy | ARC-Challenge set Accuracy | OpenBookQA Accuracy |
|---|---|---|---|---|---|
| BERT | — | — | 58.33 | 38.16 | 56.8 |
|  | — | GCN | 59.6 | 39.7 | 64.1 |
|  | ER Attention | — | 60.6 | 38.9 | 62.3 |
|  | ER Attention | GCN | **60.13** | **41.23** | **69.8** |
| XLNet | — | — | 52.53 | 40.39 | 68.8 |
|  | — | GCN | 53.7 | 41.9 | 69 |
|  | ER Attention | — | 53.5 | 41.32 | 70.1 |
|  | ER Attention | GCN | **56.56** | **43.23** | **73.40** |
| RoBERTa | — | — | 65.45 | 46.76 | 65.2 |
|  | — | GCN | 67 | 47.65 | 66.4 |
|  | ER Attention | — | 67 | 47.65 | 66.4 |
|  | ER Attention | GCN | **70.81** | **48.73** | **72.3** |
| RoBERTa[a] | — | — | 76 | 59 | 73.8 |
|  | — | GCN | 79.6 | 62.45 | 77.4 |
|  | ER Attention | — | 79.1 | 61.8 | 76.8 |
|  | ER Attention | GCN | **81** | **63** | **78** |

[a] RoBERTa-large was finetuned on the RACE (Lai et al., 2017b) dataset

Table 4: Results of various experiments, using different combinations of text encoding model, graph encoding, and external infusion using Entity-Relationship(ER) attention for the ActKnow model. All the models were trained for 10 master epochs. We achieve 2-4% improvements with inclusion of external knowledge on ARC-easy and challenge set and 5-10% increase in accuracy score on OpenBookQA dataset.

Our motivation is to show that the text-based encoding model can be improved with the introduction of graph-based encoding and external knowledge infusion.

## 6.1 Comparison with Text-based model

We compare our proposed model with various text-based ( Transformer, BERT, XLNet, and RoBERTa ) baseline models. Table 4 shows the performance of both graph-based models separately and with text-based models. These are the ablation stating the importance of the three embeddings for the ActKnow model. In all the experiments inclusion of graph-based components and external knowledge, infusion improved the results over baseline. The improvements average over 3-4% for most cases, showing that the inclusion of external knowledge and graph encoding is helpful in improving the base model. This result is obtained after training the model on the entire train set. We also find that each embedding helps the model in increasing its understanding of the language.

Here is an example from ARC-Challenge set, `Which change is the best example of a physical change? a) a cookie baking b) paper burning c) ice-cream melting d) nail rusting`, to correctly answer this question, a model must possess knowledge about physical change as well as chemical change, and also it should be able to correctly classify all the options. Our model correctly selects the options (c), but the simple text-based model selects option (b).

Table 5 demonstrates some key examples for comparison of our models with the text-based transformer pre-trained model. The first example is a simple one and can be answered correctly by all the three models i.e. RoBERTa, BaseKnow, and ActKnow, while the second question could be answered only by ActKnow. Both simple text-based models and BaseKnow marked the wrong ingredients for cooking peas. ActKnow, with better knowledge retrieval techniques, was able to have a better understanding of the overall scenario and get it correct. The fourth example is an interesting one, where all the models went wrong and put more emphasis on algae and moss, which led to marking they were producers. The sixth example represents a class of questions where incorporating knowledge did not prove to be useful and both ActKnow and BaseKnow failed. The most interesting example is the seventh one as it provides us some insight on why ActKnow works better than BaseKnow. The question was correctly answered by the language model itself, but the infusion of knowledge confused the model, and it selected the wrong option.However ActKnow training scheme reduced the importance of the knowledge infusion with the help of its entropy function, as the language model itself was quite confident. This led to the model selecting the correct answer again. This type of examples is instrumental in the performance of ActKnow. But generally, we have found that incorporating knowledge using graph-based attention methods helps, and doing so actively gives a further boost in the understanding of the model. This is particularly important in the low data regime where the number of training examples is not sufficient to get the global knowledge for solving the question. The model is not able to get complete context only from the text, but it needs real-world knowledge. Currently, for the OpenBookQA dataset, on the leader board, a variant of T5 11B (Colin Raffel, 23 Oct 2019) called UnifiedQA (Daniel Khashabi, 2 May 2020) holds the first position with an 87.2% accuracy score. The T5 based models are huge and need tons of resources for training outside our scope and hence we would like to use these models with our technique of actively infusing external knowledge in future work. Our motivation was to show that the base text encoding model could be improved with the introduction of graph-based encoding and external knowledge infusion.

| id | Example | Correct Answer | Answer Predicted By Models |
|----|---------|----------------|----------------------------|
| 1 | A positive effect of burning biofuel is | (C)powering the lights in home | (A)shortage of crops for the food supply<br>(B)an increase in air pollution<br>(C)powering the lights in a home✓✓✓<br>(D)deforestation in the amazon to make room for crops |
| 2 | Cooking peas requires | (D) turning on a stove top | (A) fresh briny sea water<br>(B) an unheated stove top<br>(C) salt and cayenne pepper✓✓<br>(D) turning on a stove top✓ |
| 3 | A rabbit may enjoy | (C)peas | (A)meat✓<br>(B)compost<br>(C)peas✓✓<br>(D)pebbles |
| 4 | shark will be unable to survive on eating algae and moss,because | (A)it is a predator | (A)it is a predator<br>(B)it is a vegetarian<br>(C)it is a freshwater fish<br>(D)it is a producer ✓✓✓ |
| 5 | Inherited behavior is exhibited when | (A)bears take a long winter sleep | (A)bears take a long winter sleep✓<br>(B)dogs sit on command<br>(C)seals clap for their trainers<br>(D)rats navigate thru a maze✓✓ |
| 6 | There is most likely going to be fog around: | (A) a marsh | (A)a marsh✓<br>(B)a tundra✓<br>(C)the plains✓<br>(D)a desert |
| 7 | f the earth was a living room, what can be done to melt the glaciers? | (A)someone would turn up the room heater | A)someone would turn up the room heater ✓✓<br>(B)someone would turn up the air conditioner<br>(C)someone would turn up the music✓<br>(D)someone would turn on the light |

Table 5: Examples demonstrating the understanding of different models.
✓ ✓ ✓ denotes the answer predicted by RoBERTa, BaseKnow and ActKnow respectively.

### 6.2 Performance Comparison in Low Data regime

Figure 5 shows the comparison of our model's performance with the text-based model concerning different low-data regime settings. The improvements average over 2-4% for most cases when using ARC easy and challenge set and 2-5% when using OpenBookQA set. Our model shows a gain of 2% inaccuracy when we use only 20% of the OpenBookQA training data, i.e., a significant improvement compared with benchmark models and text-based models in lower data regimes. The ActKnow training scheme, as shown in Figure 5 , in the lower data region brings in a further boost of 2-3%.

## 7 Ablation Study

### 7.1 Sub-Graph Size and Impact on Results

We experimented to see the effect of the size of the sub-graph generated and encoded into the model on the results. Figure 4a depicts the effect of the size of the sub-graph of the Concept-Net generated by aligning the premise-hypothesis pair on the accuracy of our model. We experimented with different sizes of the graph and for different percentages of the data. We found that the accuracy of the model went on increasing with the increase in the number of nodes up to a max value, after which it started decreasing. The reason for this may be the fact that on increasing the size of the sub-graph we include noise in the data, and the accuracy of the model suffers from that. A similar result was observed on increasing the path length as well.

### 7.2 Contribution of Each Part

To understand how each of the three embeddings ($t_i^j$, $g_i^j$, and $\mathcal{G}_i^j$) performs in the two models proposed, we carry out extensive experiments by taking different combinations of them and observing their performance for different percentages of the training data. We observed that both GCN and entity-relation embedding when added using the attention function increases the performance considerably over the text-only model. This behavior is seen for both BaseKnow and ActKnow. We further observe that combining the three embeddings using attention achieves the highest score. As the percentage of data goes on increasing, we see a crossing of the two curves representing the addition of the GCN and entity-relation respectively. At low data, the performance of the ER model is high, almost close to that of the combination, but as the percentage of the training examples increases, it is not able to keep up with its GCN counterpart which surpasses it.

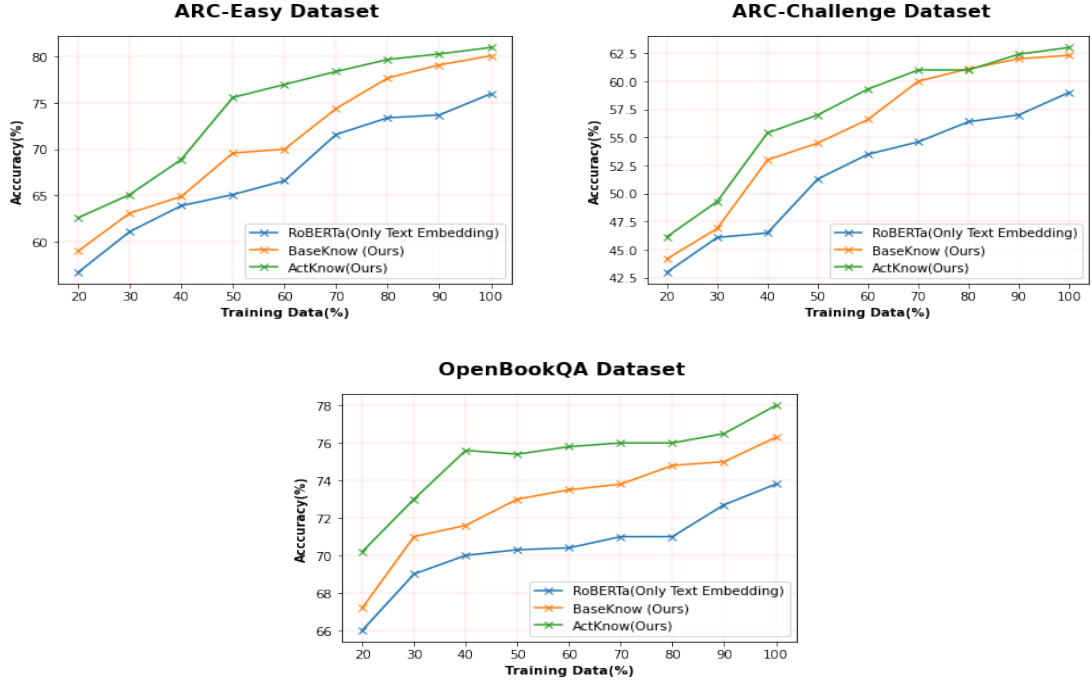

Figure 3: Performance Comparison in Low Data Regime

| Hyper-parameter | ARC | OpenBookQA |
|---|---|---|
| Model Type | Roberta-large | Roberta-large |
| Learning Rate | 1e-6 | 1e-5 |
| Max Sequence Length | 384 | 512 |
| Adam Epsilon | 1e-6 | 1e-6 |
| Adam weight decay | 0.1 | 0.1 |
| Adam ($\beta_1$, $\beta_2$) | (0.9, 0.98) | (0.9, 0.98) |
| Warm-up steps | 201 | 267 |

Table 6: Hyper-parameters which were used in experiments for ARC & OpenBookQA datasets

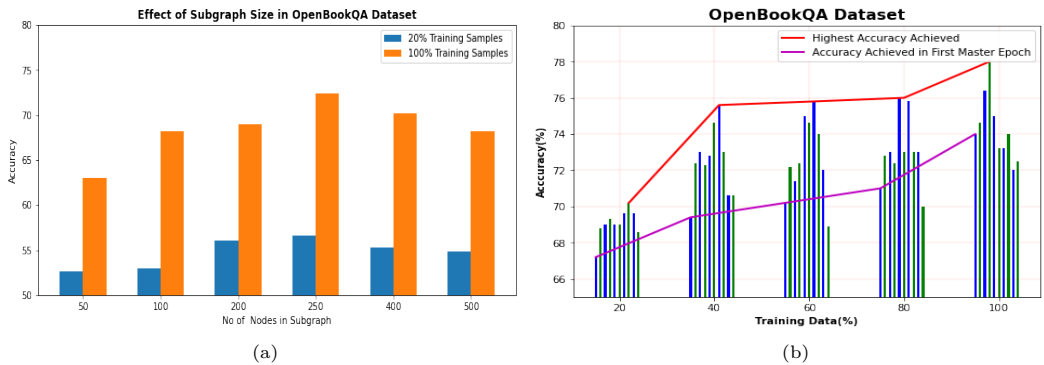

Figure 4: (a)This figure shows that on increasing the number of nodes in the sub-graph used, accuracy increases first and then decreases due to the addition of noise
(b)Figure shows the performance of every subepoch of ActKnow.The difference between the red and magenta curve shows the gain we get by iteratively training the model.

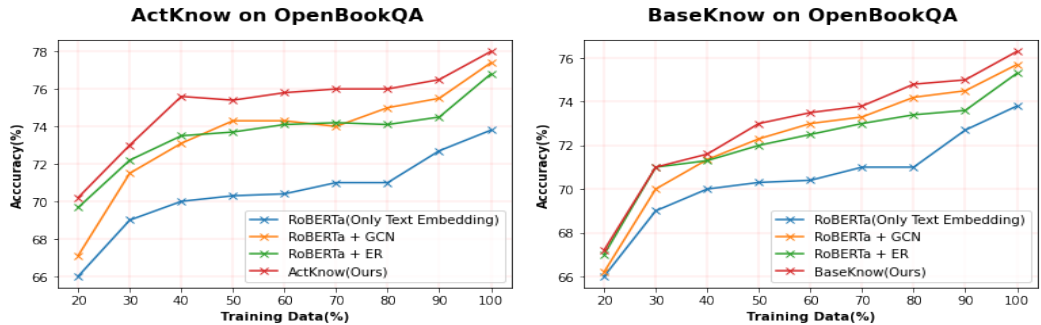

Figure 5: Ablation of different components of the model. The ER attention model performs better than the GCN model at a low percentage of training data but the curves do a crossover at a higher percentage.

### 7.3 The trend over the Master Epochs

Our model ActKnow works by changing the loss function every epoch depending on the entropy of the development set followed by fine-tuning of the model with the specific loss function. Figure 4b shows the performance of every master epoch for different percentages of data used for training. With the increase in the number of master epochs, the performance of the model goes on increasing to a point after which it starts decreasing. This point is achieved earlier for a greater percentage of the data. The figure further shows that by continuing training the model for more master epochs, we achieve a considerable improvement over the first run of the model, which can be easily seen from the gap between the two line graphs representing the performance of the first epoch and that of the epoch in which the highest accuracy was achieved. Note that a similar trend was observed on both the train and the test data.

## 8   Related Works

With the advent of Transformers (Vaswani et al., 2017; Liu et al., 2019), the state-of-the-art models for QA on various datasets are built on top of them. Pre-trained transformers bring in some general learnings that are not specific to the training data of the final task, like QA. These learnings are too diluted and do not bring in specific information required for a particular sample. We try to extract sample-specific information from external sources. Previously, external knowledge infusion is attempted in the context of NLI (Annervaz et al., 2018). Works like (Bauer et al., 2018; Mihaylov and Frank, 2018; Lin et al., 2019; Pan et al., 2019b) try to bring in external commonsense knowledge for QA. However, these works neither address the problem from a low data regime nor from on demand infusion perspective. QA being a well-studied problem, various works (Yadav et al., 2019; Banerjee and Baral, 2020; Khashabi et al., 2020) have been published recently. But these works attempt to improve the QA performance with novel models or consider other aspects of QA and, importantly do not consider the infusion of external knowledge. To the best of our knowledge, this is the first attempt to bring in active external knowledge infusion, graph-based, and text-based processing for QA, especially in low data regimes. Our experimental studies demonstrate ACTKNOW can be an essential tool in dealing with low-data regime problems in NLP and inspire future works in the space.

## 9   Conclusion & Future Work

We presented a model for the QA problem that can utilize external knowledge and complex inter-dependencies of concepts present in the problem context. Our approaches show improvement over baselines when external knowledge is incorporated. We have shown that infusing external knowledge becomes more important in the low data regime through extensive experiments. We also proposed a training method to actively seek knowledge from an external knowledge graph when the model is confused, which gave a further boost in performance. External knowledge infusion seems to be a good methodology for problems with training data scarcity. There is a multitude of options for future work. We believe that using adversarial-based training (Zhu et al., 2020) will bring more robustness to the model. In this work, we have addressed only Multiple Choice Question Answering. A similar infusion of external knowledge can be tried on other modes of Question Answering like answer extraction from text. We believe Common Sense knowledge infusion would be very effective in typical attention-based models for such problems.